\documentclass[10pt,twocolumn,letterpaper]{article}

\usepackage{cvpr}
\usepackage{times}
\usepackage{epsfig}
\usepackage{graphicx}
\usepackage{amsmath}
\usepackage{amssymb}
\usepackage{gensymb}

\usepackage{pbox}
\usepackage{booktabs}
\usepackage{pifont}

\usepackage{xcolor}
\definecolor{darkgreen}{rgb}{0.09, 0.45, 0.27}

\DeclareMathOperator*{\argmin}{argmin}

\usepackage[breaklinks=true,bookmarks=false]{hyperref}

\cvprfinalcopy % *** Uncomment this line for the final submission

\def\cvprPaperID{2954} % *** Enter the CVPR Paper ID here

\begin{document}

%%%%%%%%% TITLE
\title{Self-Improving Visual Odometry}
\vspace{-.15in}
\author{Daniel DeTone\\
Magic Leap, Inc.\\
Sunnyvale, CA\\
{\tt\small ddetone@magicleap.com}
\and
Tomasz Malisiewicz\\
Magic Leap, Inc.\\
Sunnyvale, CA\\
{\tt\small tmalisiewicz@magicleap.com}
\and
Andrew Rabinovich\\
Magic Leap, Inc.\\
Sunnyvale, CA\\
{\tt\small arabinovich@magicleap.com}
}

\maketitle
%\thispagestyle{empty}

%%%%%%%%% ABSTRACT
\begin{abstract}
\vspace{-.15in}

We propose a self-supervised learning framework that uses unlabeled monocular video sequences to generate large-scale supervision for training a Visual Odometry (VO) frontend, a network which computes pointwise data associations across images. Our self-improving method enables a VO frontend to learn over time, unlike other VO and SLAM systems which require time-consuming hand-tuning or expensive data collection to adapt to new environments. Our proposed frontend operates on monocular images and consists of a single multi-task convolutional neural network which outputs 2D keypoints locations, keypoint descriptors, and a novel point stability score. We use the output of VO to create a self-supervised dataset of point correspondences to retrain the frontend. When trained using VO at scale on 2.5 million monocular images from ScanNet, the stability classifier automatically discovers a ranking for keypoints that are not likely to help in VO, such as t-junctions across depth discontinuities, features on shadows and highlights, and dynamic objects like people. The resulting frontend outperforms both traditional methods (SIFT, ORB, AKAZE) and deep learning methods (SuperPoint and LF-Net) in a 3D-to-2D pose estimation task on ScanNet.

\end{abstract}
\vspace{-.2in}
%%%%%%%%% BODY TEXT
\section{Introduction}
\vspace{-.08in}
Simultaneous Localization and Mapping (SLAM) is an important problem in robotics, autonomous vehicles and augmented reality. Visual SLAM is one flavor of SLAM which operates on visual data, typically gray-scale or color image sequences. Visual Odometry (VO) is similar to Visual SLAM but is less complicated because it only optimizes over a recent set of observations, not requiring additional subsystems such as a re-localization and loop closure.

\begin{figure}[h]
\begin{center}
\vspace{-.14in}
\includegraphics[width=1.0\linewidth]{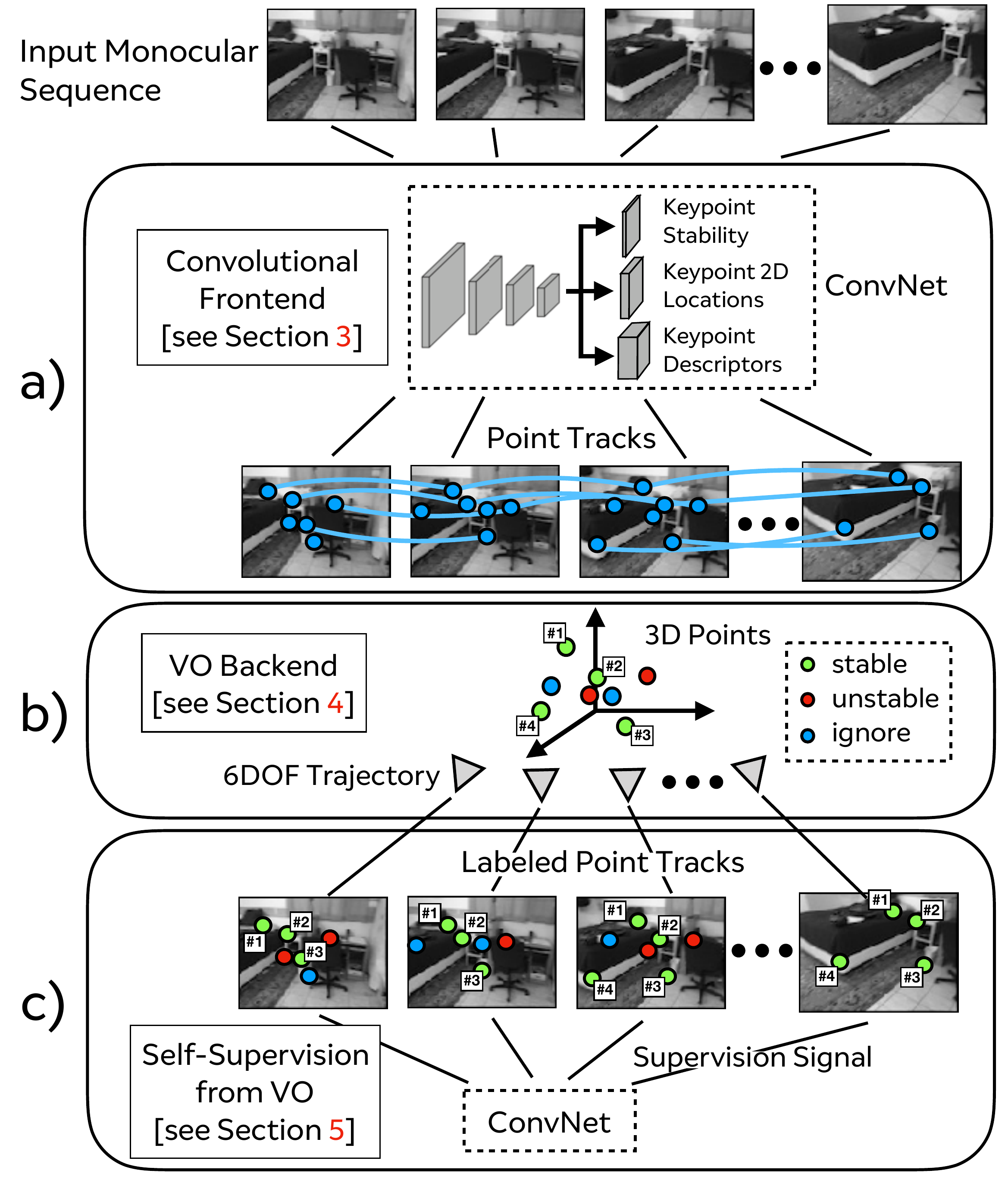}
\end{center}
\vspace{-.15in}
\caption{{\bf Self-Improving Visual Odometry.} \textbf{a)} Our Convolutional Frontend computes points and descriptors for each image in a monocular sequence to form point tracks. \textbf{b)} The VO Backend performs Bundle Adjustment to upgrade the tracks to 3D points and to classify their stability based on re-projection error. \textbf{c)} Stable 3D points act as supervision signal when we re-train the ConvNet. \label{fig:overview}} 
\vspace{-.3in}
\end{figure}

There are many excellent Visual SLAM and VO systems that exist today, but SLAM still struggles in many scenarios involving robustness, scalability and life-long operation. This is described in detail in Cadena \etal's historical overview of SLAM~\cite{cadena16}. Additionally, the constraints of augmented reality wearable devices and consumer robotics require SLAM algorithms be computationally lightweight and at the same time adaptable to new environments. To achieve this, a new type of SLAM system is required--as described in \emph{Spatial AI} by Andrew Davison \cite{davison18}--an approach which can use its own model of the world to self-supervise its ability to compute correspondence across time.

\begin{figure*}[]
\begin{center}
%\vspace{.1in}
\includegraphics[width=1.0\linewidth]{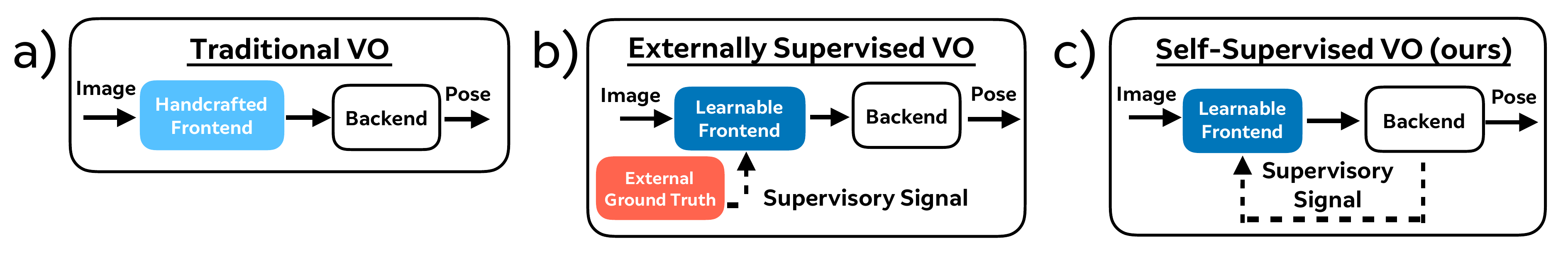}
\end{center}
\vspace{-.15in}
\caption{{\bf Visual Odometry Paradigms.} {\bf a)} A Traditional VO System uses a hand-crafted frontend combined with a backend to compute pose for an image. {\bf b)} An Externally Supervised VO system is able to improve with more data, but requires an external mechanism to provide ground-truth data, which can be expensive. {\bf c)} Self-Supervised VO uses its own outputs as supervision for the learnable frontend.  \label{fig:vo_paradigms}} 
\vspace{-.15in}
\end{figure*}

To overcome these challenges, we propose to use the temporal consistency of sequences for self-improvement. We combine a learnable frontend with a VO style optimization over sparse 2D keypoints which have been tracked over time. The key idea behind Self-Improving Visual Odometry is: the best points for a VO system are those which can be stably tracked and matched over many views. 

A visual overview of the paper is shown in Figure \ref{fig:overview}. We discuss our network architecture in Section~\ref{sec-frontend}, the backend optimization of VO in Section~\ref{sec-backend}, VO supervision in Section~\ref{sec-supervision}, pose estimation results in Section~\ref{sec-evaluation}, and conclude with discussion in Section~\ref{sec-conclusion}. The {\bf contributions} of our work are as follows: 
\begin{itemize} 
\item A novel self-labeling framework which runs VO on the outputs of a convolutional frontend and classifies their stability. The stable points are then fed back into the convolutional frontend and used as supervision to improve the system. 
\item A novel stability classifier in the frontend which predicts the stability of points from a single image, and learns to suppress points which are problematic for VO, such as points at depth discontinuities and points on dynamic objects such as people.
\end{itemize}

\section{Related Work}

{\bf Traditional VO.} The traditional approach to visual odometry~\cite{fraundorfer2012visual,nister2004visual} is based on handcrafted visual features (see Figure~\ref{fig:vo_paradigms}a). To modify or adapt most SLAM/VO systems based on handcrafted features to different sensors and environments, practitioners typically inject hand-tuned heuristics or tune hyper-parameters, often at the expense of performance in other scenarios. This happens because existing systems have little to no learned components and that ground truth data collection for SLAM/VO is very expensive and time-consuming.
 
{\bf Externally Supervised VO.} A small number of solutions in recent years such as LIFT \cite{yi16} and SuperPoint \cite{detone18} formulate the task of extracting image features and matching them across time as a learning problem and tackle them using convolutional neural networks (ConvNets). These systems can act as SLAM Frontends -- where raw images are processed and reduced to a set of geometric primitives, which are ready to be optimized by a SLAM Backend (to concurrently estimate a camera pose and 3D map). In their formulation as learning problems, these systems can improve their performance with more data, alleviating the need for heuristics. However, this is not straightforward because collecting labeled data is challenging. LIFT, for example, cleverly leverages the fact it is relatively easy to run an existing SLAM and Structure-from-Motion (SfM) systems at large-scale, since most modern systems such as ORB-SLAM \cite{mur2015} and VisualSfM \cite{wu13} can be run in real-time. LF-Net \cite{ono18} is a powerful method that uses ideas from reinforcement learning to discover keypoint locations and relies on external ground truth for camera pose and scene depths.

SIPS \cite{cieslewski18sips} uses a ranking loss to estimate a concise set of interest points. IMIPS \cite{cieslewski18imips} uses a similar approach to SIPS for learning implicit keypoint correspondence between pairs of images, without the need for descriptors. MegaDepth \cite{li18} uses SfM combined with semantic segmentation to label a large set of outdoor images. All these methods rely on external supervision which is problematic because it introduces a second SLAM system with an additional set of limitations and dependencies, which all are inherited by the learned system (see Figure~\ref{fig:vo_paradigms}b).

{\bf Self-supervised VO.} SuperPoint~\cite{detone18} takes a different approach by self-labeling a large set of images and using homographies of images to learn correspondence. While this is a surprisingly powerful technique given its simplicity, it is limited by its reliance on static images and cannot learn from real illumination changes and correspondence across difficult non-planar scenes. The backend in this method is a simple homography model, where the homographies are generated synthetically. This method falls into the self-supervised VO paradigm (see Figure~\ref{fig:vo_paradigms}c).

Geometric Matching Networks~\cite{rocco2017} and Deep Image Homography Estimation~\cite{detone16} use a similar self-supervision strategy to create training data for estimating image transformations. However, these methods lack interest points and point correspondences, which are typically required for doing higher-level computer vision tasks such as SLAM and SfM.

{\bf Learning Stability.} In \cite{zhou17}, Zhou \etal present an unsupervised approach for learning monocular depth and relative pose that does not rely on external ground truth. The model also predicts an explainability mask, which is similar to the stability classifier presented in this work because it discovers dynamic objects like people without explicitly being trained to do so. It operates on a pair of images, rather than SuperPointVO which operates on a single image.

\section{Convolutional Frontend}
\label{sec-frontend}

There are a variety of architectures available which can be trained to detect keypoints and descriptors. We choose to base the SuperPointVO architecture off of the SuperPoint~\cite{detone18} architecture because it is simple and works well in practice. We first summarize the SuperPoint architecture and describe the addition of the stability classification head; then, we describe how this architecture can be used to produce sparse optical flow tracks.

\subsection{SuperPoint Architecture Review}
The SuperPoint architecture consists of a "backbone" fully convolutional neural network, which maps the input image $I\in \mathbb{R}^{H\times W}$ to an intermediate tensor $\mathcal{B}\in \mathbb{R}^{H_c \times W_c \times F}$ with smaller spatial dimension and greater channel depth (\ie, $H_c < H$, $W_c < W$ and $F > 1$). These shared features are used in all following computation and account for the majority (roughly $90\%$) of the system compute. The computation then splits into two heads: a 2D interest point detector head and descriptor head. The interest point detector head computes $\mathcal{X}\in \mathbb{R}^{H_c\times W_c \times 65}$ and outputs a tensor sized $\mathbb{R}^{H\times W}$. The $65$ channels correspond to local, non-overlapping $8\times 8$ grid regions of pixels plus an extra ``no interest point'' dustbin. The descriptor head computes $\mathcal{D} \in \mathbb{R}^{H_c\times W_c \times D}$. We use bi-linear interpolation to sparsely upsample the descriptor field at the pixel level locations given by the interest point detector head.

In our experiments, we use the same backbone as in SuperPoint. The encoder has a VGG-like~\cite{vgg} architecture that has eight 3x3 convolution layers sized 64-64-64-64-128-128-128-128. Between every two layers there is a 2x2 max pool layer. 

\subsection{Stability Classifier Head}
The added stability classifier head operates on the intermediate features $\mathcal{B}$ output by the backbone network. It computes $\mathcal{S}\in \mathbb{R}^{H_c\times W_c \times 2}$. To compute pixel level predictions, the coarse predictions are interpolated with bi-linear interpolation and followed by channel-wise softmax over the two output channels to get the final stability probability value.
In our experiments, the stability classifier decoder head has a single 3x3 convolutional layer of 256 units followed by a 1x1 convolution layer with 2 units for the binary classification of stable versus not stable.

\begin{figure}[h]
\begin{center}
\includegraphics[width=1.0\linewidth]{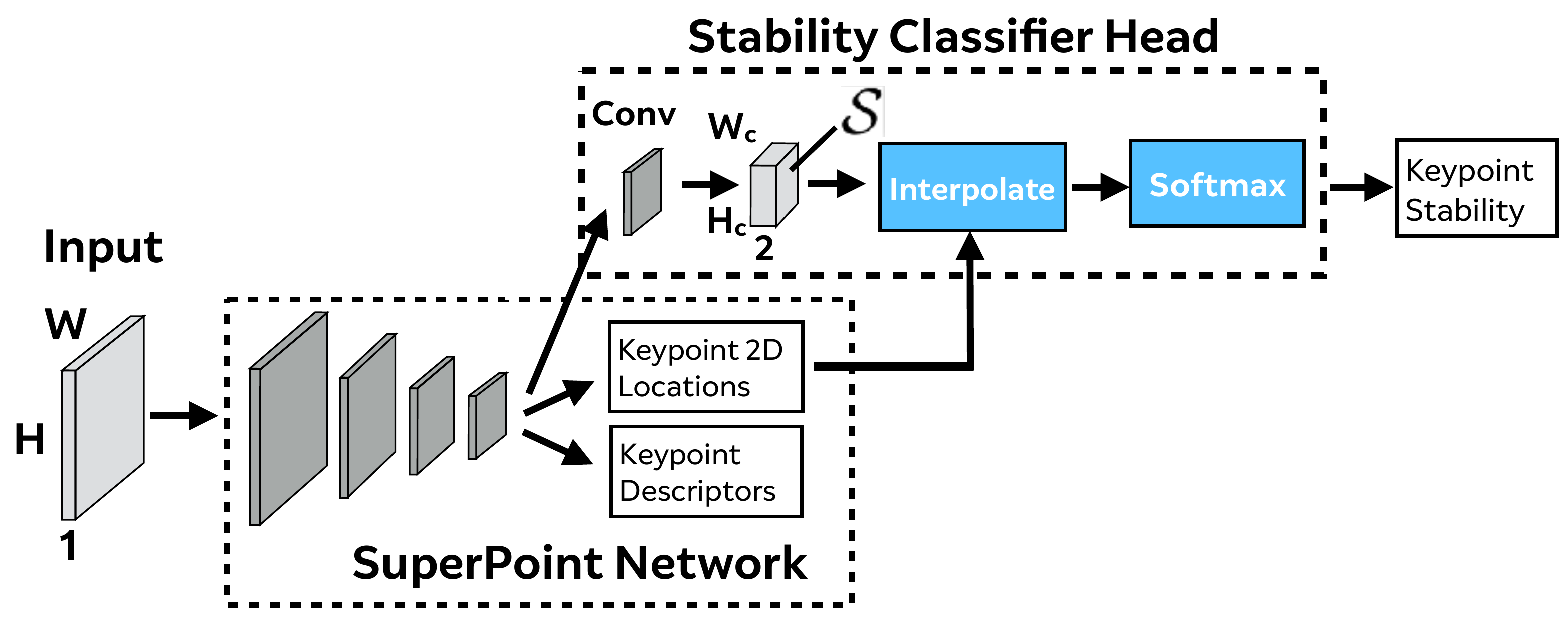}
\end{center}
\vspace{-.1in}
   \caption{\textbf{Stability Classifier Head}. To predict a stability probability for each keypoint, we augment the SuperPoint network with an additional decoder head to compute $\mathcal{S}$. }
\label{fig:stability_network}
\end{figure}

\subsection{Point Tracks}
\label{sec:point-tracks}

Once trained, the SuperPointVO Frontend can be used to form sparse optical flow tracks for an image sequence. This works by associating the points and descriptors in consecutive pairs of images with a ``connect-the-dots'' algorithm.

In other words, given a set monocular images $\mathcal{I} = \left[ I_1, I_2, \dots I_N \right]$, where $I_i \in \mathbb{R}^{H \times W}$, we can compute a corresponding set of 2D keypoints $\mathcal{U} = \left[ U_1, U_2, \dots U_N \right]$ and $U_i \in \mathbb{R}^{2 \times O_i}$ and descriptors $\mathcal{D} = \left[ D_1, D_2, \dots D_N \right]$ and $D_i \in \mathbb{R}^{256 \times O_i}$, where $O_i$ is equal to the number of points detected in the image $i$.

To match points across a pair of images $I_a$ and $I_b$, we take the bi-directional nearest neighbors of the corresponding $D_a$ and $D_b$. A bi-directional nearest neighbor match ($d_{ai}$, $d_{bj}$), where $d_{ai}, d_{bj} \in \mathbb{R}^{256}$ is one such that the nearest neighbor match from $d_{ai}$ to $D_b$ is $d_{bj}$ and the nearest neighbor match from $d_{bj}$ to $D_a$ is $d_{ai}$. This parameter-less alternative to Lowe's ratio test \cite{lowe2004} helps the algorithm use as few parameters a possible, and works well in practice. A second removal of matches is done to remove all matches such that $||d_{ai} - d_{bj}|| > \tau$ where we typically set $\tau = 0.7$. To form tracks, the same procedure is done for all consecutive pairs of images $(I_1, I_2), (I_2, I_3), \dots, (I_{N-1}, I_N)$. We found this to be a powerful heuristic in selecting good tracks, and can qualitatively be seen in Figure \ref{fig:label_examples}a.

Once the set of tracks is established, we can treat each track in the sequence as a single 3D point, and use the tracks to jointly estimate the 3D scene structure and camera poses. The following section describes this procedure.

\section{VO Backend}
\label{sec-backend}

A self-supervised Visual SLAM Frontend uses its own outputs, combined with multiple-view geometry, to create a supervised training dataset.  To achieve invariance to the non-planarity of the real world, we propose to exploit the temporal aspect of monocular video and the mostly-rigid nature of the real world. We call this extension \emph{VO Adaptation}. The key idea of VO Adaptation is to leverage VO to label which points can be stably tracked over time and use the stable tracks to learn keypoint correspondence over many views.

To describe our VO backend and thus the VO Adaptation process, we summarize some multiple-view geometry concepts in following sections to establish our mathematical notation. We refer the reader to the Hartley and Zisserman Multiple View Geometry \cite{hartley2003multiple} textbook for more details.

\vspace{-.1in}
\subsection{Optimization Variables}
\vspace{-.05in}
In a monocular sequence of $N$ images, the set of \textbf{camera poses} for the $i$-th camera are represented by their rotation and translation ($R_i$,$t_i$), where $R_i \in SO(3)$ and $t_i \in \mathbb{R}^3$.

For a scene with $M$ \textbf{3D points} which re-project into some or all of the $N$ images, each point is represented by $X_j$, where $X_j \in \mathbb{R}^3$. There is no 3D prior structure imposed on the reconstruction, other than the depth regularization function $d(Z)$ (introduced later) which penalizes point configurations too close (or behind) or too far from the camera.

The \textbf{camera intrinsics} $K$ is an upper-triangular matrix made up of focal lengths $f_x$ and $f_y$ together with the principal point ($c_x$,$c_y$). While it is possible to optimize over one $K$ for each image (as is typically done in a SfM pipeline), our VO backend assumes a single, fixed $K$.
\vspace{-.05in}
\subsection{Observation Variables}
\vspace{-.05in}
$\mathcal{U}$ is the set of \textbf{2D point observations}, a collection of N matrices, 
one for each image. $\mathcal{U} = \left[ U_1, U_2, \dots U_N \right]$ and $U_i \in \mathbb{R}^{2 \times O_i}$, where $O_i$ is equal to the number of 2D observations in the image $i$. A single image measurement is represented by $u_{ij} \in \mathbb{R}^2$.

$\mathcal{W}$ is the set of \textbf{observation confidence weights}. The observation confidence weights are used during optimization to prioritize more confidence observations over less confident ones. Each image has a set of associated scalar weights $\mathcal{W} = \left[ W_1, W_2, \dots W_N \right]$ where $W_i \in \mathbb{R}^{O_i}$. Each scalar weight ranges between zero and one, i.e.\ $w_{ij} \in [0,1]$.

$\mathcal{A}$ is the set of 3D-to-2D \textbf{association tracks}. Since every 3D point $X_j$ in the sparse 3D map is not observed in every frame due to the moving camera and scene occlusions, we have a set of 3D-to-2D association vectors for each image $\mathcal{A} = \left[ A_1, A_2, \dots A_N \right]$, where $A_i \in \mathbb{Z}^{O_i}$. Each association integer indicates the 3D map point index it corresponds to and ranges between zero and the total number of points in the scene, i.e.\ $a_{ij} \in [1,M]$.

\begin{figure*}[]
\begin{center}
\vspace{-.2in}
\includegraphics[width=1.0\linewidth]{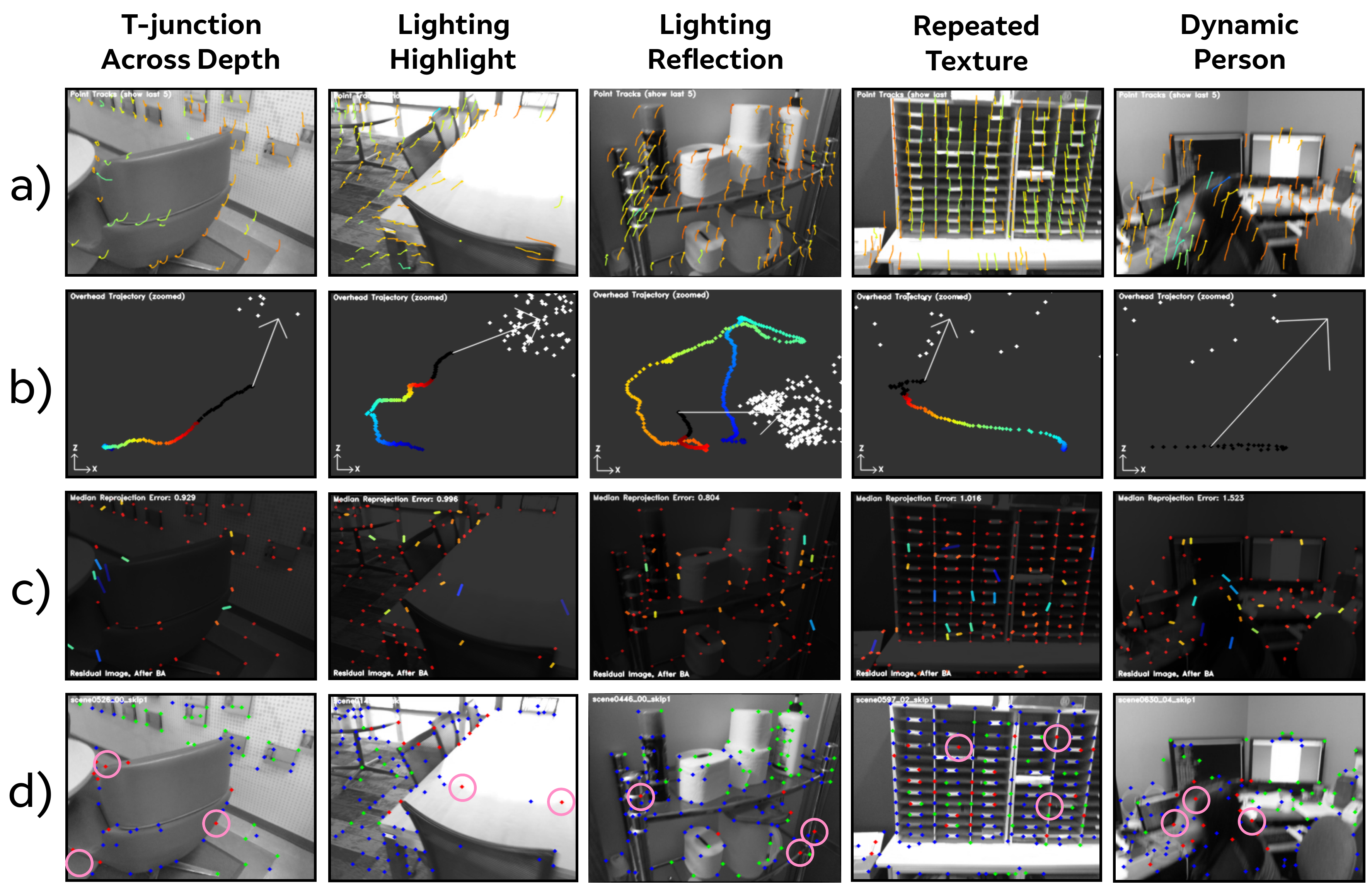}
\end{center}
\vspace{-.1in}
   \caption{\textbf{Labeling points with VO}. Five examples of patterns labeled by VO to have a low stability due to five different effects are shown in each column. We show \textbf{a)} Sparse point tracks from the convolutional frontend, \textbf{b)} Overhead projection of the computed VO backend camera pose trajectory and sparse 3D map, \textbf{c)} Re-projection error residual images (red = low error, blue = high error), and \textbf{d)} Labeled point tracks with stability labels ({\color{darkgreen}green = stable}, {\color{red}red = unstable}, {\color{blue}blue = ignore}, pink circle = characteristic example of unstable point).}
\label{fig:label_examples}
\vspace{-.1in}
\end{figure*}

\subsection{3D Projection Model}
\vspace{-.05in}
We use a pinhole camera model for camera projection which explains how a 3D world point gets projected into a 2D image given the camera pose and the camera intrinsics. Letting $X_j\in \mathbb{R}^3$ denote the $j$-th 3D point, $(R_i,t_i)$ the $i$-th camera pose, $K$ the camera intrinsics, and $u_{ij}\in\mathbb{R}^2$ the corresponding 2D projection:

\vspace{-.05in}
\begin{equation}
\begin{bmatrix}
           u_{ij1} \\
           u_{ij2} \\
           1
         \end{bmatrix} \sim K [R_i|t_i] \begin{bmatrix}
           X_{j} \\
           1
         \end{bmatrix}. 
\label{eqn:projection}
\end{equation}

The $\sim$ in Equation~\ref{eqn:projection} denotes projective equality. To simplify our calculations, we use a $\mathbb{R}^3 \rightarrow \mathbb{R}^2$ projection function $\Pi(X)$ which performs the 3D to 2D conversion,

\vspace{-.05in}
\begin{eqnarray}
\Pi(\begin{bmatrix}
           X \\
           Y \\
           Z
         \end{bmatrix}) &=& \frac{1}{Z}\begin{bmatrix}
           X \\
           Y 
         \end{bmatrix}.
\end{eqnarray}

To measure the quality of the estimated camera poses and 3D points, we can measure the re-projection of each 3D point into each camera. We write the squared \textbf{re-projection error} $e_{ij}^2$ for the $j$-th 3D point in the $i$-th image as follows:

\vspace{-.05in}
\begin{equation}
e^2_{ij} = ||\Pi(K(R_iX_{a_{ij}}+t_i))-u_{ij}||^2.
\label{eqn:reprojection}
\end{equation}

\subsection{Depth Regularization}
\vspace{-.05in}
We introduce a \textbf{depth regularization} function $d(Z'_{ij})$, where $Z'_{ij} = \left[ R_i X_{ij} + t_{ij} \right]_3$, where $\left[\cdot\right]$ means taking the third component of the vector, which incurs a quadratic penalty for estimated 3D point depths $Z'_{ij}$ which are too close or too far from the camera, parameterized by two scalars $d_{min}$ and $d_{max}$. It also prevents depths from moving behind the camera center. We found $d_{min} = 0.1$ and $d_{max} = 5.0$ to work well for indoor scenes. The term is:
\vspace{-.05in}
\begin{equation}
\begin{split}
d(Z'_{ij}) = &\max(0, Z'_{ij} - d_{max})^2 + \\ &\min(Z'_{ij}-d_{min}, 0)^2.
\end{split}
\label{eqn:depth_regularization}
\end{equation}

\subsection{Camera Pose and Point Depth Initialization}
\vspace{-.05in}
We initialize each new camera pose ($R_{N+1}$,$t_{N+1}$) with the camera pose from the previous frame ($R_{N}$,$t_{N}$). We initialize new 3D point depths to $1.0$. While it is common for traditional SfM pipeline will initialize the 3D points depths $Z'_{ij}$ using linear triangulation methods, we found that this did not improve our VO results significantly and added more complexity to the system. We found that simply initializing the point depths to unity depth and adding the depth regularization term was enough of a prior for the BA optimization to work well.

\subsection{Final Bundle Adjustment Objective}
\vspace{-.05in}
The final bundle adjustment objective is the combination of the re-projection error function $e_{ij}^2$, the depth regularization function, the 2D observation weights $w_{ij}$ and a Huber robust loss function $\rho(\cdot)$ to help deal with outliers. We denote the final objective function for BA, $\Omega_{BA}(\cdot)$, as follows:

\vspace{-.2in}
\begin{equation}
\begin{split}
\Omega (\cdot) = \sum_{i=1}^{N} \sum_{j=1}^{O_i} w_{ij} \rho \left(e_{ij}^2 + d(Z'_{ij})\right).
\end{split}
\label{eqn:ba-obj}
\end{equation}

% \begin{empheq}[box={\mymath[colback=red!30,drop lifted shadow, sharp corners]}]{equation}
\vspace{-.05in}
\begin{equation}
\{R^*,t^*\}_{i=1}^N,X^* = \argmin_{\{R,t\}_{i=1}^N,X} \Omega(\{R,t\}_{i=1}^N,X|K,\mathcal{U},\mathcal{W},\mathcal{A})
\label{eqn:ba-final}
\end{equation}

\subsection{VO Backend Implementation}
The BA optimization is done over a fixed window of the most recent $N_{last} = 30$ poses, corresponding to about one second of motion. We use the ceres-solver \cite{ceres} c++ package to perform the Levenberg-Marquardt optimization over Equation \ref{eqn:ba-final}. For each new image, we run BA for up to 100 iterations, which on average takes about one second per frame. Row b in Figure \ref{fig:label_examples} shows some example trajectories in a jet colormap, where blue is the earlier part of the trajectory, red is the latter part of the trajectory, and black is the $N_{last}$ poses, while the white points show the 3D points $X_j$.

\section{Self-Supervision from VO}
\label{sec-supervision}
We combine the SuperPoint-based VO Frontend described in Section \ref{sec-frontend} with the VO Backend system described in Section \ref{sec-backend} to run on monocular video sequences.

\subsection{Labeling Stability}
Once VO is complete for a given sequence, the number of observations and re-projection errors for each 3D point are used to label stability. If a point is tracked for a reasonably long time, we can use its reprojection error to classify it as stable versus non-stable. Let $T_j$ denote the number of observations tracked to form a 3D point $X_j$. Let $\mathrm{mean}({e_j})$ and $\mathrm{max}({e_j})$ be the mean and maximum of the re-projections respectively into each observed camera. We define the stability $S_j$ of that 3D point as:

\begin{equation}
S_{j}= 
\begin{cases}
    \mathrm{{\color{darkgreen}stable}},& \text{if $\left(T_j \geq 10\right)$ and $\left(\mathrm{mean}({e_j}) \leq 1 \right)$}  \\
    \mathrm{{\color{red}unstable}},& \text{else if $\left(T_j \geq 10\right)$ and $\left(\mathrm{max}({e_j}) \geq 5 \right)$} \\
    \mathrm{{\color{blue}ignore}},              & \text{otherwise}
\end{cases}
\label{eqn:stability}
\end{equation}

In other words, stable points are those which have been tracked for at least ten frames and have an average re-projection error less than one pixel. Unstable points are those which have been tracked for at least ten frames and have a maximum re-projection error of more than five pixels. The points which do not satisfy these two constraints are ignored during training--the network can decide to treat them as stable or unstable as it chooses.

\begin{figure}[ht!]
\begin{center}
\includegraphics[width=1.0\linewidth]{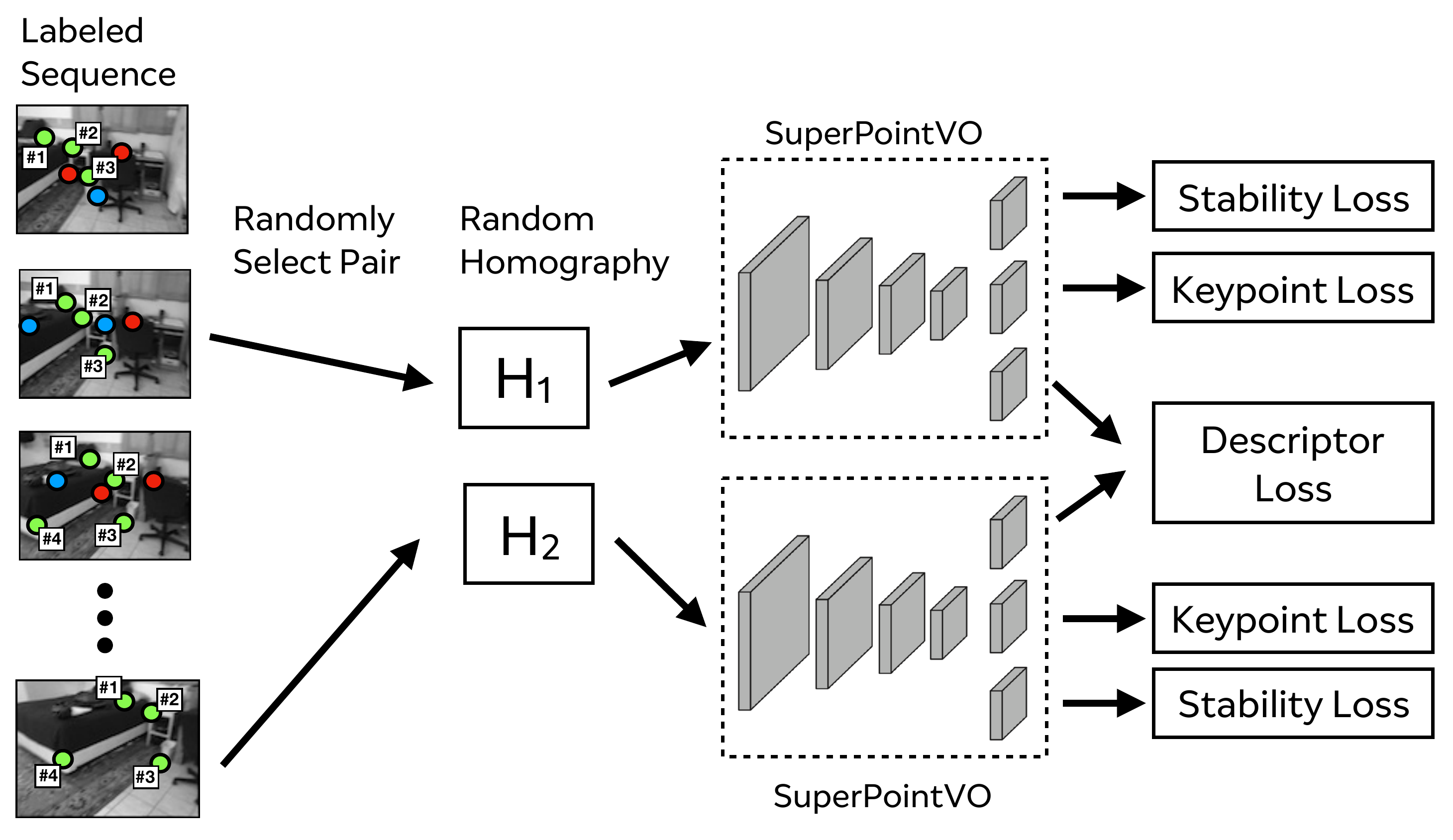}
\end{center}
\vspace{-.1in}
\caption{{\bf Siamese Training Set-up.} The SuperPointVO network is trained using Siamese learning. Random nearby image pairs are selected from a labeled sequence, and warped by a random homography before being used to train the three network tasks.  \label{fig:training}} 
\vspace{-.05in}
\end{figure}

The self-labeling procedure discovers unstable regions such as t-junctions across depth discontinuities, features on shadows and highlights, and dynamic objects like people. See examples of this in Figure \ref{fig:label_examples} row d.

\subsection{Training Details}

SuperPointVO is trained on the full ScanNet \cite{dai17} training set, consisting of 2.5 million images collected from 1500 diverse indoor scenes. VO Adaptation requires only monocular images and intrinsics calibration information, so the depth frames and pose information is not used during training time.

The network is trained by loading random nearby pairs from sequences which have been labeled by the VO backend as described in Section \ref{sec-backend}. The Siamese training set-up is pictured in Figure \ref{fig:training}. We follow the same methodology of SuperPoint~\cite{detone18}, where the descriptor is trained using Siamese metric learning, and the keypoint detector is trained using a standard softmax + cross entropy loss. The pairs are randomly sampled from a temporal window of +/- 60 frames, resulting in pairs with a maximum time window of about 4 seconds. The loss functions also incorporate the ``ignore class,'' which is used for unknown correspondences and unknown 2D point locations.

To train the stability classifier, we add an extra loss term to the final loss of SuperPoint $\mathcal{L}_s$ which denotes the stability loss. The stability loss is trained with a standard binary cross-entropy loss function.

\section{Evaluation}
\label{sec-evaluation}
 
\subsection{3D-to-2D Pose Estimation}
We compare SuperPointVO to various Frontend systems in their ability to do 3D-to-2D pose estimation, a critical component of Visual Odometry. To evaluate this, we follow the 3D-to-2D methodology described in Scaramuzza and Fraundorfer~\cite{scaramuzza11}. We first load random pairs of images from a given sequence, separated by 30 frames, or about 1 second worth of motion. Correspondences are chosen using matches which are the nearest neighbor to each other, as described in Section \ref{sec:point-tracks}, except with no $\tau$ match score threshold, to keep the comparison fair among all methods. Next, rather than triangulating the point depths from 2D-to-2D correspondences, we use the depth from the RGBD frame of one of the images in the image pair. This simulates a sparse 3D map built from integration and estimation of many previous views. We then use the default OpenCV \texttt{SolvePnPRansac()} function to estimate the camera pose given 3D-to-2D matches.

\begin{table}[h]
  \centering
  \def\arraystretch{1}
  \setlength{\tabcolsep}{2pt}
  \begin{tabular}{c|ccc|ccc}
          \toprule
          Frame Difference & 30 & 60 & 90 & 30 & 60 & 90  \\
          \midrule
          Average Relative Pose & 15\degree & 28\degree & 38\degree  & 25cm  & 48cm & 67cm \\
          \midrule
          & \multicolumn{3}{c|}{Rot. Error $<$ 5\degree} & \multicolumn{3}{c}{Transl. Error $<$ 5cm}  \\
          \midrule
          ORB                  & .432 & .154 & .074 & .285 & .076 & .036 \\
          AKAZE                & .641 & .238 & .120 & .413 & .114 & .056\\
          SIFT                 & .650 & .325 & .181 & .448 & .156 & .083\\
          SURF                 & .698 & .322 & .172 & .457 & .152 & .069 \\
          LF-Net               & .803 & .425 & .233 & .524  & .194 & .094 \\
          SuperPoint (COCO)    & .818 & .488 & .283 & .587  & .250 & .116 \\
          SuperPoint (ScanNet) & .836 & .499 & .288 & .613  & .267 & .128 \\
          SuperPointVO (ours)  & \textbf{.848} & \textbf{.536} & \textbf{.331}  & \textbf{.622}  & \textbf{.277} & \textbf{.140} \\
          \bottomrule
  \end{tabular}
  \smallskip
  \caption{\textbf{ScanNet Pose Estimation Accuracy}. Best results are marked in bold.  }
  \label{tbl:pose-errors}
\end{table}

SuperPointVO is compared to full sparse feature matching pipelines: SuperPoint, LF-NET, SIFT\cite{lowe2004}, SURF\cite{bay08}, AKAZE \cite{alcantarilla12} and ORB \cite{rublee2011}, which each computes 2D keypoints with corresponding descriptors. We allow each method to detect up to 500 keypoints per frame. For SuperPoint, we compare to both the author's original implementation which was trained on MS-COCO images and a variant we trained from scratch on ScanNet. For LF-Net, we use the authors' implementation, trained on the ScanNet training set. Note that LF-Net requires both the depths and camera poses for training, while our method does not. We use the default OpenCV implementation for all other methods. The systems are tested on the full ScanNet test set, consisting of 100 diverse indoor scenes. For each scene, 50 random pairs are selected, resulting in 5000 pairs. We experiment with different frame differences: 30, 60, and 90 corresponding to one, two, and three-second intervals from the 30 FPS camera. Results are presented in Table \ref{tbl:pose-errors}. SuperPointVO achieves the best pose estimation accuracy across all frame differences. 

\begin{figure*}[]
\begin{center}
\vspace{-.04in}
\includegraphics[width=.95\linewidth]{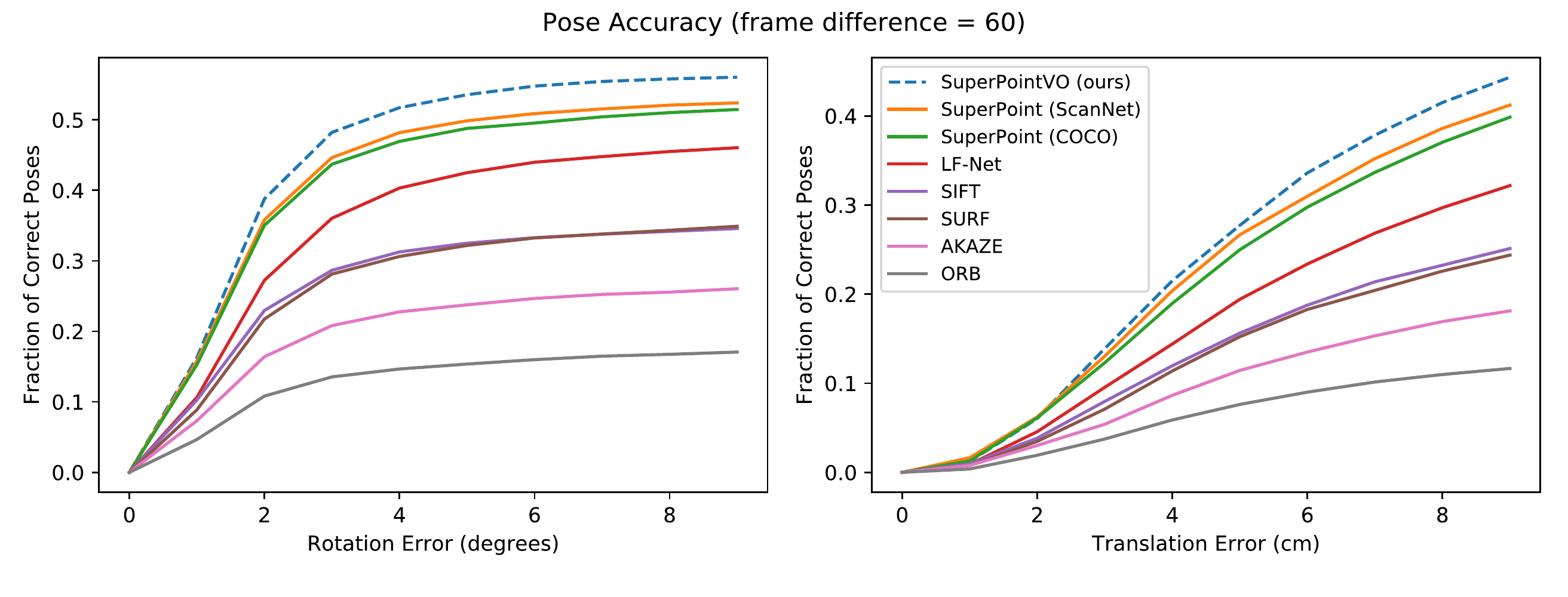}
\end{center}
\vspace{-.15in}
\caption{{\bf 3D Pose Estimation Comparison.} Pose estimation accuracy on the ScanNet indoors dataset with a frame difference of 60. Our SuperPointVO method outperforms all other methods at various pose thresholds. \label{fig:scannet_errors}} 
\vspace{-.1in}
\end{figure*}

The benefit of SuperPointVO over SuperPoint trained on ScanNet is the most apparent at larger baselines. For example, there are relative improvements of 1.5\%, 7.5\%, 15\% for rotation error estimation at 30, 60 and 90 frame difference settings respectively. This makes sense -- the homography assumption made in SuperPoint breaks down more at larger camera baselines in non-planar scenes.

\begin{figure*}[]
\begin{center}
\includegraphics[width=1.0\linewidth]{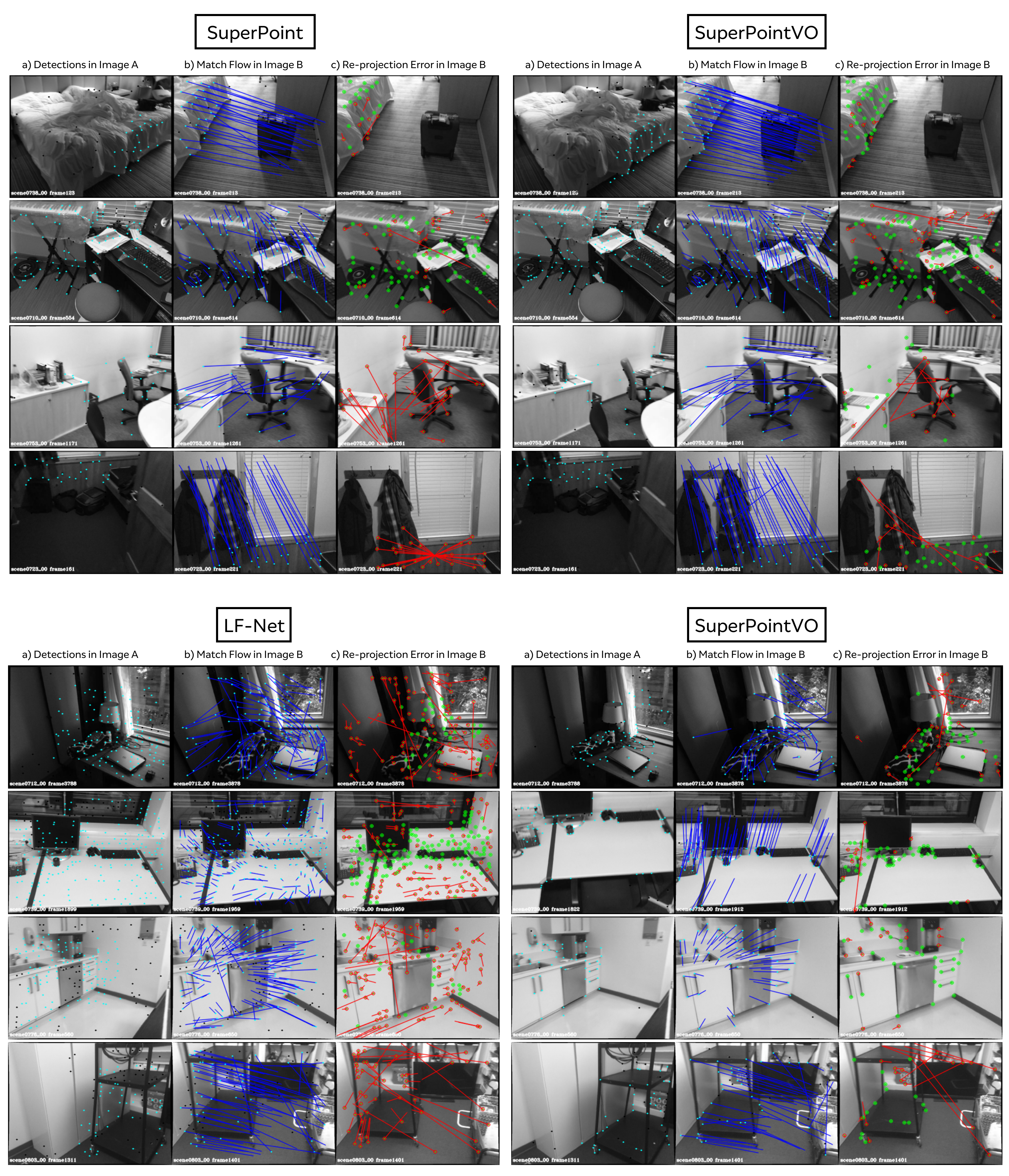}
\end{center}
\vspace{-.1in}
   \caption{\textbf{Qualitative ScanNet Comparison}. SuperPointVO is compared to both SuperPoint and LF-Net. Each row represents a different example. The triplets of images are explained as follows: \textbf{a)} Detections in Image A are shown in cyan. \textbf{b)} Detections in Image B are shown in cyan, with the match flow vectors drawn in blue. \textbf{c)} After the relative pose is estimated via PnP + RANSAC, the points from Image A are transformed to Image B using the estimated pose. If the re-projection error is less than 3 pixels, the projection is colored green--otherwise it is colored red, and the re-projection residual vector is drawn. Best viewed in color.}
\label{fig:scannet_qualitative}
\vspace{-.1in}
\end{figure*}

Qualitatively, when compared to SuperPoint, our SuperPointVO variant is better at wide-baseline matching, especially in non-planar scenes. When compared to LF-NET, SuperPointVO detect fewer points in texture-less regions, which can be problematic for LF-Net in pose estimation. See Figure~\ref{fig:scannet_qualitative} for examples.

\subsection{VO Trajectory Evaluation}
\vspace{-.05in}
 To evaluate the quality of the stability classifier outputs, we run two variants of SuperPointVO, one with the stability classifier head enabled and another with it disabled. We follow the evaluation protocol of \cite{zhang18}, where relative rotation and translation errors are computed for varying sub-trajectory lengths. To make the VO estimation difficult, we run VO with every 10th frame as input. We exclude trajectories with any invalid ground truth pose, which results in 57/100 valid trajectories. The stability confidence values for each keypoint are used in the backend optimization by setting the value of $w_{ij}$ for each keypoint. When the stability is disabled, $w_{ij}=1.0$. The addition of stability into VO improves trajectory estimation, as shown in Table ~\ref{tbl:vo-errors}.

\begin{table}[h]
  \centering
  \small
  \def\arraystretch{1}
  \setlength{\tabcolsep}{2pt}
  \begin{tabular}{c|ccc|ccc}
          \toprule
          Sub-Trajectory Length & 2 sec & 5 sec & 10 sec & 2 sec & 5 sec & 10 sec  \\
          \midrule
          & \multicolumn{3}{c|}{Rot. Error (\degree)} & \multicolumn{3}{c}{Transl. Error (cm)}  \\
          \midrule
          No Prediction              & 22.9 & 49.1 & 70.9 & 39.6 & 80.3 & 116 \\
          \midrule
          SuperPointVO (-stability)  & 2.09 & 4.45 & 7.89 & 5.5 & 14.0 & 32.0 \\
          SuperPointVO (+stability)  & \textbf{1.93} & \textbf{4.12} & \textbf{7.26}  & \textbf{5.1}  & \textbf{12.9} & \textbf{26.5} \\
          \bottomrule
  \end{tabular}
  \smallskip
  \caption{\textbf{VO Trajectory Evaluation}. Using the stability classifier during VO improves performance. Best results are marked in bold.  }
  \label{tbl:vo-errors}

\end{table}

\vspace{-.05in}
\subsection{Visualizing Stability}
\vspace{-.05in}

\begin{figure}[h]
\begin{center}
\includegraphics[width=1.0\linewidth]{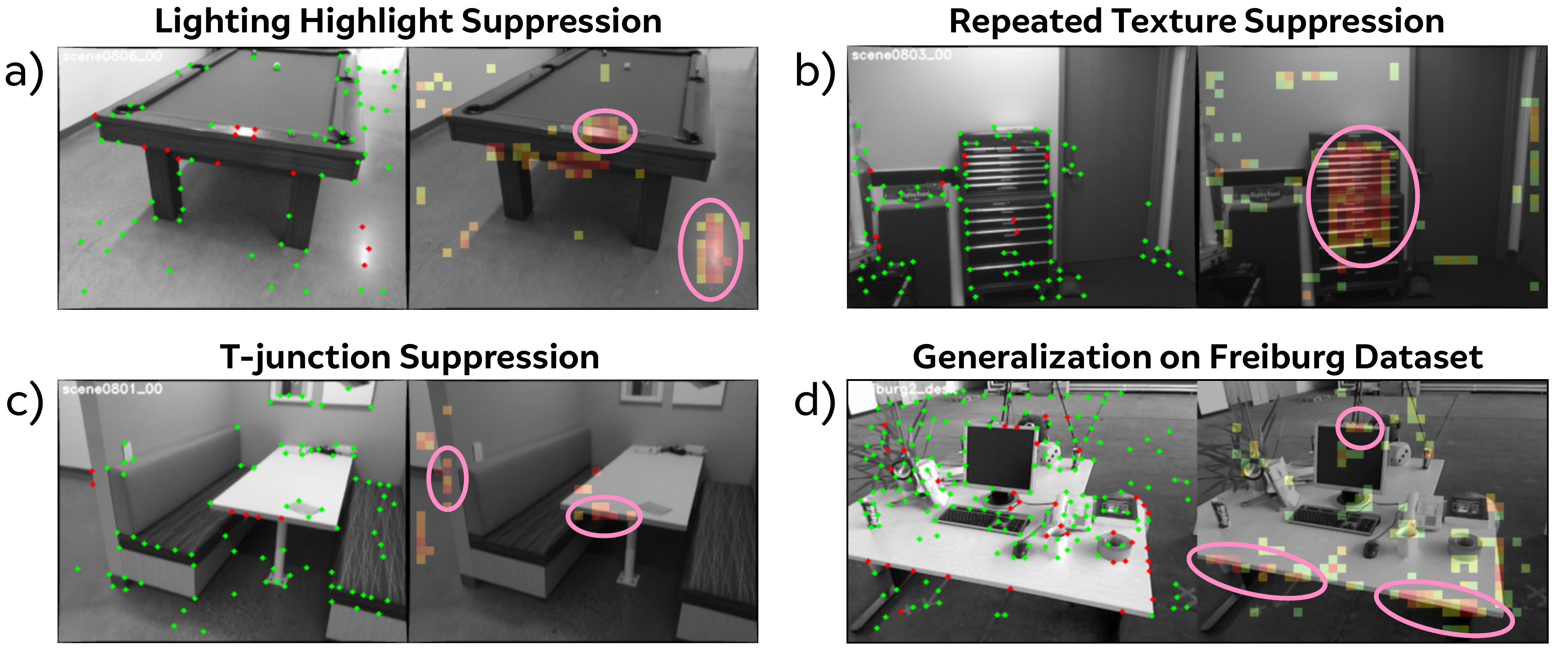}
\end{center}
\vspace{-.1in}
\caption{{\bf Stability Classifier in Action.} Sample visualizations of dense per-pixel stability predictions depicting three types of ``low stability'' regions such as {\bf a)} lighting highlights, {\bf b)} repeated texture, and {\bf c)} t-junctions. {\bf d)} shows an example on a different dataset. The pink circles highlight suppressed regions. \label{fig:stability_qualitative}} 
\vspace{-.1in}
\end{figure}

When trained at scale, the stability classifier discovers regions of the image which are likely to result in unstable tracks (i.e., large reprojection error) during VO. Even though the stability classifier's training data comes from VO tracks (see Figure~\ref{fig:label_examples}d), we can apply the resulting stability classifier to create dense stability heatmaps as shown in Figure \ref{fig:stability_qualitative}. The most common types of regions that our method deems unstable are the following: t-junctions, lighting highlights, and repeated texture. We lastly show an example in Figure \ref{fig:stability_qualitative}d of stability on the Freiburg RGBD dataset \cite{sturm12iros}.

\section{Conclusion}
\label{sec-conclusion}
In this paper, we presented a self-supervised method for improving convolutional neural network VO frontends. The approach works by combining an existing frontend with a traditional VO backend to track points over time and estimate their stability via a re-projection error metric. Our method does not rely on expensive data collection and uses the VO output for self-supervision. When trained on a monocular video dataset comprising 2.5 million images, the resulting system out-performs existing methods (both traditional and learning-based) for the task of pose estimation, in both small and wide-baseline settings. The system automatically learns which points are good for VO and suppresses unstable points, such as those caused by lighting highlights and from dynamic objects.

\clearpage

{\normalsize%\small
\bibliographystyle{ieee}
\bibliography{main}
}

\end{document}